\documentclass{article}
\usepackage{ijcai21}

\usepackage{times}
\usepackage{soul}
\usepackage{url}
\usepackage[hidelinks]{hyperref}
\usepackage[utf8]{inputenc}
\usepackage[small]{caption}
\usepackage{graphicx}
\usepackage{amsmath}
\usepackage{amsthm}
\usepackage{booktabs}
\usepackage{algorithm}
\usepackage{algorithmic}
\usepackage{mathtools}
\usepackage{resizegather}
\usepackage{dsfont}
\usepackage{enumitem}
\usepackage{caption}
\usepackage{multirow}
\usepackage{amsfonts}
\usepackage{amssymb}
\usepackage{amsthm}
\usepackage{amsmath}
\usepackage{float}
\usepackage[font={rm,md,up}]{subfig}
\usepackage{ragged2e}
\usepackage{tikz}
\usetikzlibrary{positioning}
\usetikzlibrary{arrows.meta}
\usepackage{multirow}
\usepackage{arydshln}

\tikzset{
  >={Latex[width=1mm,length=2mm]}
}

\tikzset{every picture/.style={line width=0.75pt}} 

\usepackage{epsfig}

\title{Domain Generalization under Conditional and Label Shifts via Variational Bayesian Inference}

\author{
Xiaofeng Liu$^{1,2}$ 
Bo Hu$^{2,3}$ 
Linghao Jin$^{2,4}$ 
Xu Han$^{2,4}$ 
Fangxu Xing$^{1}$ 
Jinsong Ouyang$^1$ 
Jun Lu$^2$ 
Georges El Fakhri$^{1}$ 
Jonghye Woo$^1$\\
\affiliations
$^1$Dept. of Radiology, Massachusetts General Hospital and Harvard Medical School, Boston, MA, USA\\
$^2$Beth Israel Deaconess Medical Center and Harvard Medical School, Boston, MA, USA\\
$^3$National University of Singapore, Singapore\\
$^4$Johns Hopkins University, Baltimore, MD, USA\\
\emails
liuxiaofengcmu@gmail.com, huboniccolo@bupt.edu.cn, ljin23@jhu.edu, xhan32@jhu.edu, jlu@bidmc.harvard.edu, \{fxing1, ouyang.jinsong, elfakhri, jwoo\}@mgh.harvard.edu
}

\begin{document}

\maketitle
\begin{abstract}
In this work, we propose a domain generalization (DG) approach to learn on several labeled source domains and transfer knowledge to a target domain that is inaccessible in training. Considering the inherent conditional and label shifts, we would expect the alignment of $p(x|y)$ and $p(y)$. However, the widely used domain invariant feature learning (IFL) methods relies on aligning the marginal concept shift w.r.t. $p(x)$, which rests on an unrealistic assumption that $p(y)$ is invariant across domains. We thereby propose a novel variational Bayesian inference framework to enforce the conditional distribution alignment w.r.t. $p(x|y)$ via the prior distribution matching in a latent space, which also takes the marginal label shift w.r.t. $p(y)$ into consideration with the posterior alignment. Extensive experiments on various benchmarks demonstrate that our framework is robust to the label shift and the cross-domain accuracy is significantly improved, thereby achieving superior performance over the conventional IFL counterparts.


\end{abstract}

\section{Introduction}

Deep learning usually relies on the independent, identically distributed (i.i.d.) assumption of training and testing datasets, while target tasks are usually significantly heterogeneous and diverse \cite{che2019deep,liu2021mutual}. This motivates many researchers to investigate domain adaptation (DA) \cite{liu2021subtype} and domain generalization (DG) \cite{matsuura2019domain}. The source and target domain shifts are expected to be compensated for by a variety of adaptation steps. Despite the success of DA in several tasks, much of the prior work relies on utilizing the massive labeled/unlabeled target samples for its training. 

The recently aroused DG task \cite{matsuura2019domain} assumes that several labeled source domains are available in training without access to the target sample/label. Collectively exploiting these source domains can potentially lead to a trained system that can be generalized well on a target domain \cite{hu2019domain}. A predominant stream in DG is the domain invariant feature learning (IFL) \cite{ghifary2016scatter}, which attempts to enforce $p^i(f(x))=p^j(f(x))$, where $i$ and $j$ index the two different source domains. The typical solution can be obtained via momentum or adversarial training \cite{ghifary2016scatter}.

\begin{figure}[t]
\centering
\includegraphics[width=8.5cm]{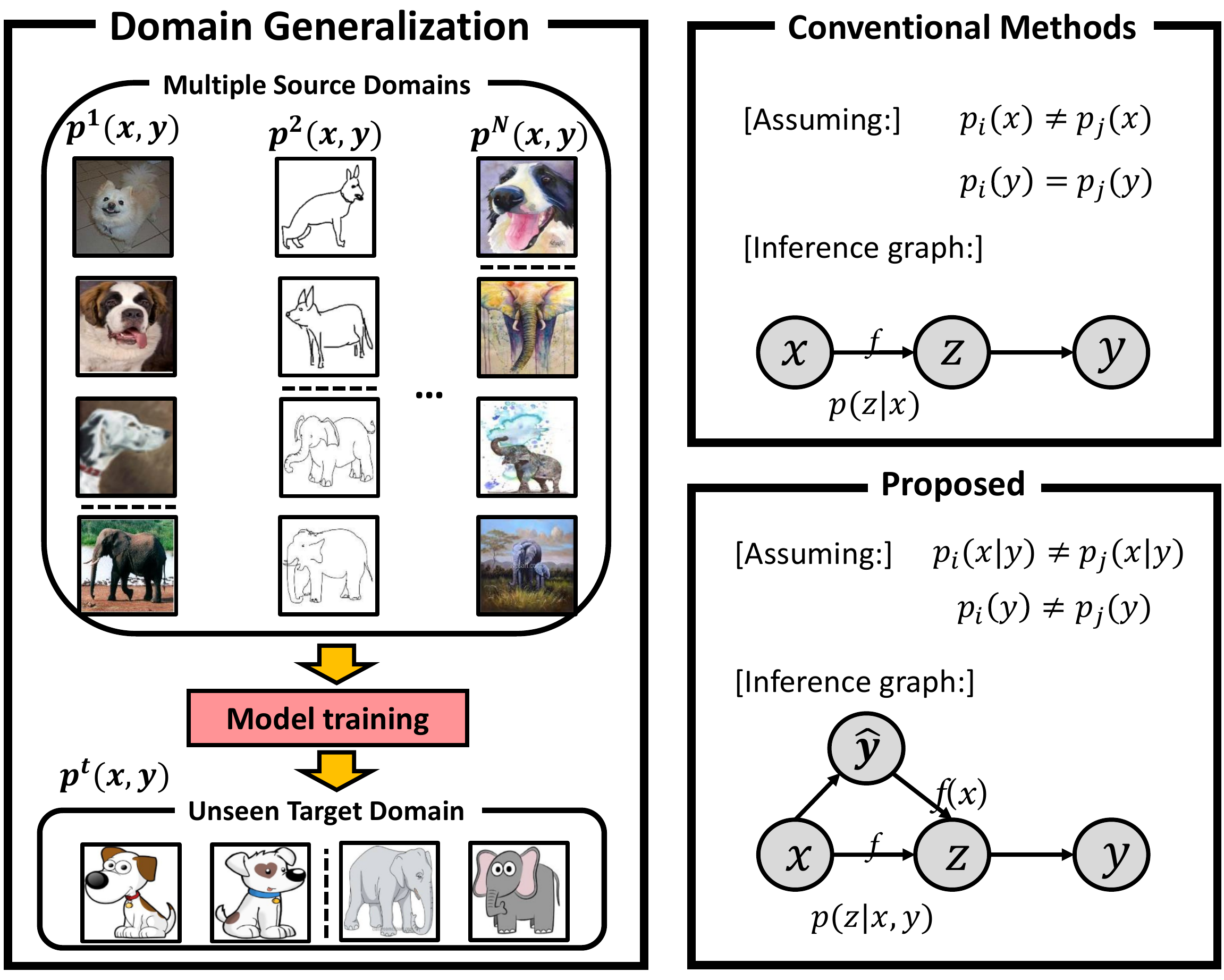}\\ 
\caption{Illustration of the DG and the comparison of coventional IFL and our solution w.r.t. the presumption and inference graph.}\label{fig:11} 
\end{figure}

However, the different classes can follow different domain shift protocols, e.g., the street lamp will be sparkly at night, while the pedestrian is shrouded in darkness. Therefore, we would expect the fine-grained class-wise alignment of the condition shift w.r.t. $p(f(x)|y)$, where $f(\cdot)$ is a feature extractor \cite{zhang2013domain,gong2016domain,combes2020domain}.

Assuming there is no concept shift (i.e., $ p^i(y|f(x))=p^j(y|f(x))$) and label shift (i.e., $p^i(y)=p^j(y))$), and given the Bayes' theorem, $p(f(x)|y)=\frac{p(y|f(x))p(f(x))}{p(y)}$, IFL is able to align $p(f(x)|y)$ if $p^i(f(x))=p^j(f(x))$.


However, the label shift $p^i(y)\neq p^j(y)$ \cite{liu2021subtype}, i.e., class imbalance, is quite common in DG, as illustrated in Fig. \ref{fig:11}. Since $f(\cdot)$ is a deterministic mapping function, IFL is able to encode the domain invariant representation under the $covariate$ $shift$ assumption (i.e., only $p^i(x)\neq p^j(x)$) \cite{moreno2012unifying}. Under the label shift, the $covariate$ alignment cannot be used as an alternative of $conditional$ alignment (i.e., $p^i(x|y)=p^j(x|y))$ \cite{li2018deepcc}. Actually, both the conditional and label shifts are the realistic setting in most of DG tasks.

Recently, \cite{li2018deepcc} proposes to align the conditional shift, assuming that there is no label shift. However, it is ill-posed to only consider one of conditional or label shift \cite{zhang2013domain,kouw2018introduction}. To mitigate this, both the $conditional$ and $label$ shifts are taken into account for DA from a causal interpretation view \cite{zhang2013domain,gong2016domain}. However, its linearity assumption might be too restrictive for real-world challenges.

In this work, we first analyze the different shift conditions in real-world DG, and rethink the limitation of conventional IFL under different shift assumptions. Targeting the conditional and label shifts, we propose to explicitly align $p(f(x)|y)$ and $p(y)$ via variational Bayesian inference and posterior label alignment. Note that the fine-grained class-wise $p(f(x)|y)$ alignment can lead to $p(f(x))$ alignment following the law of total probability \cite{zhang2013domain}.

Aligning the conditional distribution $p(f(x)|y)$ across source domains under the label shift is usually intractable. Thus, we propose to infer the domain-specific variational approximations of these conditional distributions, and reduce the divergence among these approximations. 
   
Specifically, we enforce the conditional domain invariance by optimizing two objectives. The first one enforces the approximate conditional distributions indistinguishable across domains by matching their reparameterized formulations (i.e., the mean and variance of Gaussian distribution). The second one maximizes the probability of observing the input $x$ given the latent representation and domain label, which is achieved by a domain-wise likelihood learning network. Assuming that the conditional shift is aligned, we can then align the posterior classifier with the label distribution following a plug-and-play manner.


The main contributions are summarized as follows:

$\bullet$ We explore both the conditional and label shifts in various DG tasks, and investigate the limitation of conventional IFL methods under different shift assumptions.

$\bullet$ We propose a practical and scalable method to align the conditional shift via the variational Bayesian inference.

$\bullet$ The label shift can be explicitly aligned by the posterior alignment operation. 

We empirically validate its effectiveness and generality of our framework on multiple challenging benchmarks with different backbone models and demonstrate superior performance over the comparison methods.

\begin{figure}[t]
\centering
\includegraphics[width=8cm]{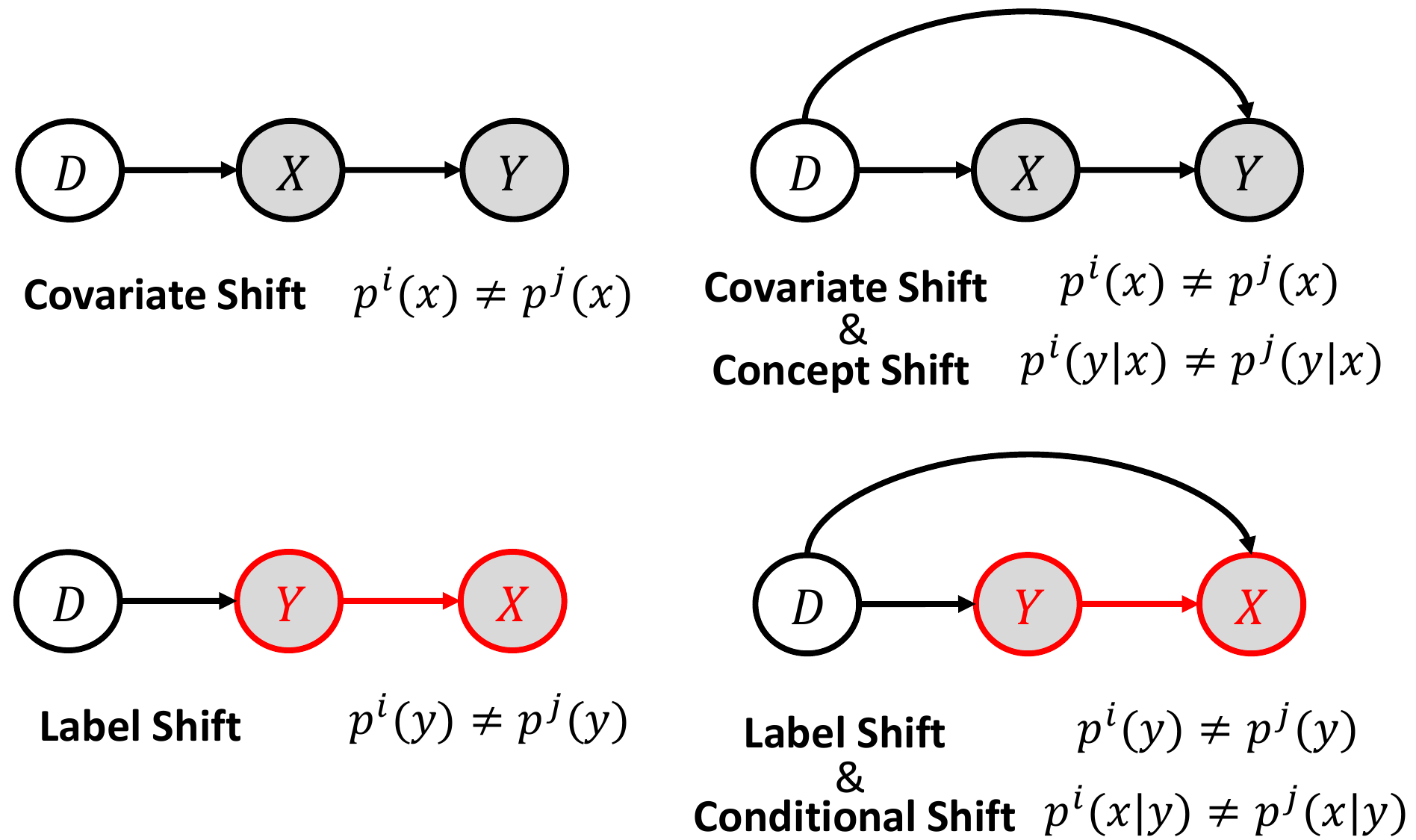} 
\caption{Summary of the possible shifts/combinations, and their inherent casual relationships. We target the last one, and $p(x|y)$ is the fine-grained class-wise distribution of $p(x)$ [Zhang $et~al.,$ 2013]. The observed variables are shaded. The red circle is used to emphasize the $y$ to $x$ causality.
}\label{fig:22} 
\end{figure}

\section{Related Work}

DG assumes that we do not have prior knowledge about the target domain in that we do not have access to labeled/unlabeled samples of the target domain at the training stage \cite{matsuura2019domain}. The conventional DG methods can be divided into two categories. The first strategy aims to extract the domain invariant features with IFL \cite{ghifary2016scatter}. A typical solution is based on adversarial learning, which reduces the inter-domain divergence between the feature representation distributions \cite{li2018domain,liu2021mutual}.  

The other strategy focuses on the fusion of domain-specific feature representations (DFR-fusion). \cite{mancini2018best} builds the domain-specific classifiers by multiple independent convolutional networks. Then, it uses a domain agnostic component to fuse the probabilities of a target sample belonging to different source domains. \cite{ding2017deep} infers the domain-invariant feature by matching its low-rank structure with domain-specific features. Typically, these DG methods assume that $p(y)$ is invariant across domains. Therefore, aligning $p(f(x))$ can be a good alternative to align the conditional shift. However, this assumption is often violated due to the label shift in real-world applications. Therefore, independent conditional and label shift assumptions are more realistic in real-world applications.

Domain shifts in DA can be categorized into covariant, label, conditional, and concept shifts \cite{moreno2012unifying,zhang2015multi}. In this work, we examine these concepts and adapt their causal relationships to DG, as summarized in Fig. \ref{fig:22}. Conventionally, each shift is studied independently, by assuming that the other shifts are invariant \cite{kouw2018introduction}. For example, \cite{li2018deepcc} aligns the conditional shift assuming that no label shift occurs. We note that the concept shift usually has not been considered in DG tasks, since most of the prior work assumes that an object has different labels in different domains. Some recent works~\cite{zhang2013domain,gong2016domain,combes2020domain} assume that both conditional and label shifts exist in DA tasks and tackle the problem with a causal inference framework. However, its linearity assumption and location-scale transform are too restrictive to be applied in many real-world applications. It is worth noting that under the conditional and label shift assumption, $y$ is the cause of $x$, and therefore it is natural to infer $p(x|y)$ of different domains directly as in \cite{scholkopf2012causal,gong2018causal} as a likelihood maximization network.

In this work, we propose a novel inference graph as shown in Fig. \ref{fig:11} to explicitly incorporate the conditional dependence, which is trained via variational Bayesian inference.

\section{Methodology}

We denote the input sample, class label, and domain spaces as $\mathcal{X}, \mathcal{Y}$, and $\mathcal{D}$, respectively. With random variables $x\in\mathcal{X}$, $y\in\mathcal{Y}$, and $d\in\mathcal{D}$, we can define the probability distribution of each domain as $p(x,y|d)$. For the sake of simplicity, we assume $y$ and $d$ are the discrete variables for which $\mathcal{Y}=\left\{1,2,\dots,K\right\}$ is the set of class labels. In DG, we are given $N$ source domains $\{p^i(x,y)\}_{i=1}^N$ to train the latent representation encoder $f(x)$ and the representation classifier $p(y|f(x))$ \cite{li2018domain}. The trained and fixed encoder and classifier are used to predict the labels of samples drawn from the marginal distribution $\{p^t(x)\}$ of an ``unseen" target domain $\{p^t(x,y)\}$.

The conventional IFL assumes that $p(y)$ and $p(y|x)$ are invariant across domains. Since $f(\cdot)$ is a deterministic mapping function, $p(y|f(x))$ should also be invariant across domains. Therefore, if $p(f(x))$ of different domains are aligned, the conditional shift, $p(f(x)|y)=\frac{p(y|f(x))p(f(x))}{p(y)}$, is also aligned. However, with the conditional and label shift assumption, the alignment of $p(f(x)|y)$ and $p(y)$ is more challenging than the covariant shift which only requires to align the marginal distribution $p(f(x))$.

We note that with the law of total probability, $p(f(x))$ can be aligned if the fine-grained $p(f(x)|y)$ is aligned \cite{zhang2013domain}. Moreover, $p(y)$ for all source domains can be calculated by simply using the class label proportion in each domain. Besides, modeling $p(f(x)|y)$ is natural, since it follows the inherent causal relation under the conditional and label shift (see Fig. \ref{fig:22}).

For the simplicity and consistency with the notation of autoencoder, we denote $f(x)$ as $z$\footnote{We use $f(x)$ and $z$ interlaced to align with the conventional IFL and variational autoencoder literature, respectively.}, which is the latent variable encoded from $x$ by $f$. We note that $z=f(x)$ is dependent on its corresponding input sample $x$. The class conditional distribution $p(f(x)|y)$ can be reformulated as $p(z|x,y)$.

\begin{figure}[t]
\centering
\includegraphics[width=8.5cm]{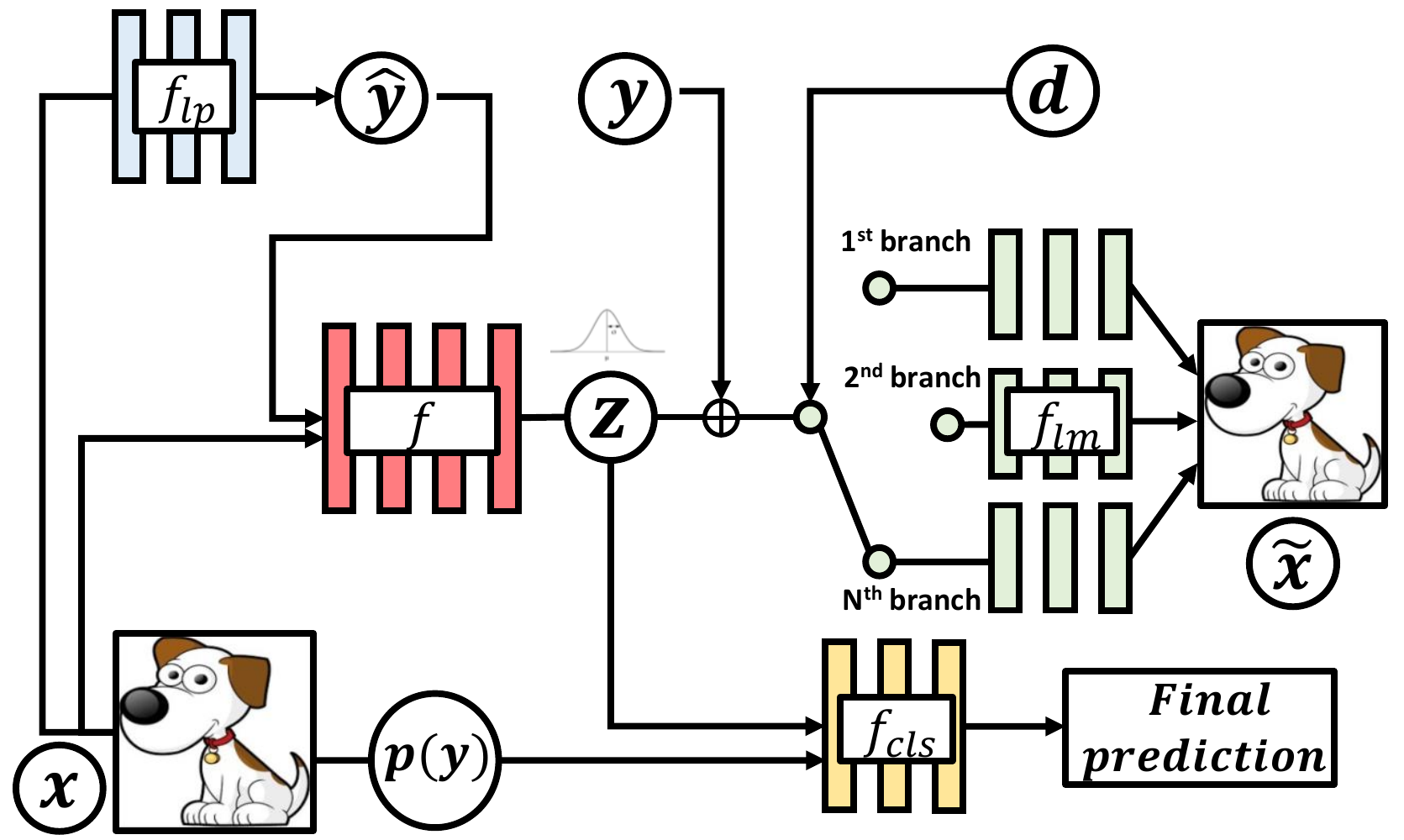}\\ 
\caption{Illustration of our framework, which consists of a conditional feature encoder $f$, a domain-wise likelihood maximization network $f_{lm}$, a label prior network $f_{lp}$, and a classifier $f_{cls}$. Our latent representation $z$ is aligned with the Gaussian prior. $f$, $f_{lp}$, $f_{cls}$ are used in testing, $f_{lm}$ is only for training.}\label{fig:33} 
\end{figure}

The corresponding inference graph and detailed framework are shown in Fig. \ref{fig:11} and Fig. \ref{fig:33}, respectively. When inferring the latent representation $z$, we explicitly concatenate $x$ and $y$ as input. Moreover, the final class prediction is made by a posterior alignment of the label shift, which also depends on label distribution $p(y)$.

\subsection{Variational Bayesian Conditional Alignment} 

Although $p(f(x))$ and $p(y)$ for all source domains can be modeled by IFL and class label proportion, respectively, $p(z|x,y)$ is usually intractable for moderately complicated likelihood functions, e.g., neural networks with nonlinear hidden layers. While this could be solved by the Markov chain Monte Carlo simulation (MCMC), this requires expensive iterative inference schemes per data point and does not scalable to the large-scale high-dimensional data. 

To align the class-dependent $p(z|x,y)$ across different domains, we first construct its approximate Gaussian distribution $q(z|x,y)$, and resort to the variational Bayesian inference \cite{kingma2013auto} to bridge it with a simple Gaussian family for which the inference is tractable. Specifically, we have the following proposition:\vspace{+5pt}

\noindent\textbf{Proposition 1.} The minimization of the inter-domain conditional shift $p(z|x,y)$ is achieved when its approximate distribution $q(z|x,y)$ is invariant across domains, and the KL divergence between $p(z|x,y)$ and $q(z|x,y)$ is minimized.\vspace{+5pt}

\noindent\textbf{Proof 1.} A simple justification can be: 

\noindent if we have $~KL(q_1(z|x,y)||p_1(z|x,y))=0$,\\ $~~~~~~~~~~~~~~~~~~~~~~KL(q_2(z|x,y)||p_2(z|x,y))=0$,\\  $~~~~~~~~~~~~~~~~~~~~~~KL(q_1(z|x,y)||q_2(z|x,y))=0$,\\ then, we have $KL(p_1(z|x,y)||p_2(z|x,y))=0$.

Following the variational bound \cite{kingma2013auto}, minimizing the KL divergence between $p(z|x,y)$ and $q(z|x,y)$ is equivalent to maximizing the evidence lower bound (ELBO) \cite{domke2018importance} of the likelihood $\text{log}p(x,y,z)$, denoted by $\mathcal{L}$:\begin{equation}
\begin{aligned}
   {max}~ \mathcal{L} =  {min}~ D_{KL}(q(z|x,y)||p(z|x,y)),    \label{eq:1}
\end{aligned}\end{equation} where the $D_{KL}$ term is the KL divergence of the approximate from the true posterior and the ELBO of the likelihood $\text{log}p(x,y,z)$, i.e., $\mathcal{L}$ can be re-written as \begin{equation}
\begin{aligned}
   \mathcal{L} =   \int q(z|x,y) \text{log}  \frac{p(x,y,z)}{q(z|x,y)}dz,   \label{eq:2}
\end{aligned}\end{equation} which can be reformulated as \begin{equation}
\begin{aligned}
   \mathcal{L} = - D_{KL}(q(z|x,y)||p(z|y))+\mathbb{E}_{z\sim q(z|x,y)} [\text{log}p(x|y,z)], \label{eq:3}
\end{aligned}\end{equation}where $\mathbb{E}$ denotes the expectation. Therefore, approximating $p(z|x,y)$ with $q(z|x,y)$ requires two objectives, i.e., minimizing $D_{KL}(q(z|x,y)||p(z|y))$, while maximizing the expectation of $\text{log}p(x|y,z)$. 

For $N$ domains, $p^i(z|y), i\in\{1,2,\dots,N\}$ is the prior distribution in the variational model, e.g., multivariate Gaussian distribution. When $p^i(z|y)$ is sampled from the same Gaussian distribution and is invariant across the source domains, the first objective in Eq.~(\ref{eq:3}), i.e., $D_{KL}(q(z|x,y)||p(z|y))$, can explicitly enforce $q^i(z|x,y)$ to be invariant across domains.

By further incorporating the second objective into Eq.~(\ref{eq:3}), we attempt to minimize the KL  divergence of $p(z|x,y)$ and $q(z|x,y)$ as in Eq.~(\ref{eq:1}). Then, $p^i(z|x,y)$ should be invariant across the source domains, i.e., 
\begin{equation}
\begin{aligned}
p^1(z|x,y)=p^2(z|x,y)=\cdots =p^N(z|x,y)
\end{aligned} 
\end{equation}

Actually, if we have $p_1(y|x)=p_1(x|y)p_1(y)/p_1(x)$, $p_2(y|x)=p_2(x|y)p_2(y)/p_2(x)$, and $p_1(x|y)=p_2(x|y)$, $p_1(x)=p_2(x)$, then $p_1(y|x)/p_1(y)= p_2(y|x)/p_2(y)$.

The first optimization objective of Eq.~(\ref{eq:3}) targets to align the conditional distribution $q(z|x,y)$ across the source domains. Since the prior distribution $p(z|y)$ is the multivariate Gaussian  distribution, it is also natural to configure $q(z|x,y)$ as multivariate Gaussian. Practically, we follow the reparametric trick of variational autoencoder \cite{kingma2013auto} in such a way that the inference model has two outputs, i.e., $\mu_j^i$ and $\sigma_j^i$ which are the mean and variance of the Gaussian distribution $q(z_j^i|x_j^i,y_j^i)$. Then, $z_j^i=\mu_j^i+\sigma_j^i\odot\epsilon$, where $\epsilon\in N(0,I)$. Without loss generality, we have: \begin{equation}
\begin{aligned}
  {L_1} = \sum_{i=1}^{N} \sum_{j=1}^{M_i}  \left[-\text{log}\frac{\sigma_j^i}{\bar{\sigma} }+\frac{\bar{\sigma}^2+(\bar{\mu}-\mu_j^i)^2}{2(\sigma_j^i)^2}-\frac{1}{2}\right],\label{eq:6666}
\end{aligned}\end{equation} where $M_i$ is the number of input in a batch from domain $i$. Usually, we set the prior $p^i(z|y)$ to be the standard multivariate Gaussian distribution, where the mean and variance are $\bar{\mu}=0$ and $\bar{\sigma}=1$, respectively.

The second optimization objective of Eq. (\ref{eq:3}) aims to maximize the probability of observing the input $x$ given $y$ and $z$. We propose to configure a likelihood maximization network $f_{lm}$. It maximizes the likelihood that the latent feature representation of images in a specific domain can effectively represent the same-class images in this domain. Practically, our $f_{lm}$ contains $N$ sub-branches, each of which corresponds to a domain. At the training stage, we choose the corresponding branch according to the source domain label $d$. Its loss can be formulated as\begin{equation}
\begin{aligned}
   {L_2} = \sum_{i=1}^{N} \sum_{j=1}^{M_i} ||f_{lm}^i(z_j^i,y_j^i)-x_j^i||^2,
\end{aligned}\end{equation} which solves the maximum likelihood problem by minimizing the difference between the input data and the generated data in the corresponding domain. We note that $f_{lm}$ is only used in training. 

Posterior collapses is a long-lasting problem of VAEs using continuous Gaussian prior. The recent progress of discrete priors or flow model can be potentially added on our model. However, our training does not suffer from it significantly. Note that each of the $N$ decoders is only trained with about $1/N$ data of encoder $f$ and it is likely that the weak decoder can also be helpful. Using an approximation with multivariate Gaussian prior offers much better tractability as in variational autoencoder (VAE), which is good for posterior modeling.

\subsection{Label-prior}

Inferencing the latent representation $z$ requires to know the label information $y$ in advance, since we are modeling the approximate conditional distribution $q(z|x,y)$. Although $y$ is always available in training, the ground truth $y$ is not available in testing. We note that $f_{lm}$ is only used in training which always has $y$.

To alleviate this limitation, we infer the label from the input image $x$ as a prior to control the behavior of the conditional distribution matching module. Specifically, we configure a label-prior network $f_{lp}:x\rightarrow \hat{y}$ to infer the pseudo-label, and use $\hat{y}$ as input to the posterior label alignment classifier $f_{cls}$ in both training and testing. Our label prior network $f_{lp}$ is trained by the cross-entropy loss $L_{CE1}$ with the ground-truth label $y$.  

Moreover, at the training stage, we can further utilize the ground-truth $y$ and minimizing $D_{KL}(q(z|x,y)||q(z|x,\hat{y}))$ to update the encoder $f$. We denote the to be minimized KL divergence as $L_{\hat{y}}$. Minimizing $L_{\hat{y}}$ is not mandatory, while $L_{\hat{y}}$ can encourage the encoder to be familiar with the noisy $\hat{y}$ \cite{lian2019learning} and learn to compensate for the noisy prediction. We note that assigning $\hat{y}$ to an uniform histogram as the dialog system \cite{lian2019learning} to fill the missing variate $y$ can degenerate the modeling of $p(f(x)|y)$ to $p(f(x))$ in our DG task. Therefore, the pseudo-label will be post-processed by both encoder and classifier, which may adjust the unreliable $\hat{y}$.

\begin{algorithm}[t]
\caption{VB inference under conditional and label shifts}
\label{alg:A}
\begin{algorithmic}
\STATE {Initialize network parameters} 
\REPEAT 
\STATE//Construct the mini-batch for the training 
\STATE$\left\{x,y,d\right\}\leftarrow$ random sampling from the dataset
\STATE//Forward pass to compute the inference output
\STATE$\hat{y}\leftarrow f_{lp}(x)$; $z\leftarrow f(x,\hat{y})$;
\STATE$\tilde{x}\leftarrow f_{lm}(z,\hat{y})$; ${y}\leftarrow f_{cls}(z,y)$
\STATE//Calculate the loss functions: $L_1$, $L_2$, $L_{CE1}$, $L_{CE2}$
\STATE//Update parameters according to gradients
\STATE $f\leftarrow L_1+\alpha L_2+\beta L_{CE2}+\theta L_{\hat{y}}$;
\STATE $f_{lp}\leftarrow L_{CE1}$;  
\STATE $f_{lm}\leftarrow L_2$; 
\STATE {$f_{cls}\leftarrow L_{CE2}$}
\UNTIL{deadline}
\end{algorithmic}
\end{algorithm}

\subsection{Posterior Alignment with Label Shift}

Finally, we align $p(y)$ to obtain the final classifier $f_{cls}$. Since the classifier is deployed on all of the source domains, we can regard all of the source domains as a single domain, and denote the classifier as \begin{equation}
\begin{aligned}
\bar{p}(y|f(x))=\frac{\bar{p}(f(x)|y)\bar{p}(y)}{\bar{p}(f(x))},
\end{aligned}
\end{equation}
where $\bar{p}(f(x)|y)$, ${\bar{p}(f(x))}$, and ${\bar{p}(y)}$ are its class-conditional, latent representation, and label distribution, respectively.  

Suppose that all the conditional distribution ${p}^i(f(x)|y)$ and the latent distribution $p^i(f(x))$ are aligned to each other using variational Bayesian inference, they should also be aligned with $\bar{p}(f(x)|y)$ and ${\bar{p}(f(x))}$. Therefore, the posterior alignment and the final prediction of the sample $f(x)$ from domain $i$ can be formulated as \begin{equation}
\begin{aligned}
   {p}^i(y=k|f(x))=\frac{{p}^i(y=k)}{\bar{p}(y=k)}\bar{p}(y=k|f(x)),
\end{aligned}
\end{equation} where $\bar{p}(y=k|f(x))$ is the $k$-{th} element value of the classifier's softmax prediction. Here, we also calculate the cross entropy loss $L_{CE2}$ between $p^i(y=k|f(x))$ and the ground truth label $y$.

As detailed in Algorithm \ref{alg:A}, we update $f$ with $L_1+\alpha L_2+\beta L_{CE2}+\theta L_{\hat{y}}$, update $f_{lp}$ with $L_{CE1}$, update $f_{lm}$ with $L_2$, and update $f_{cls}$ with $L_{CE2}$, respectively.




\begin{figure}[t]
\centering
\includegraphics[width=7cm]{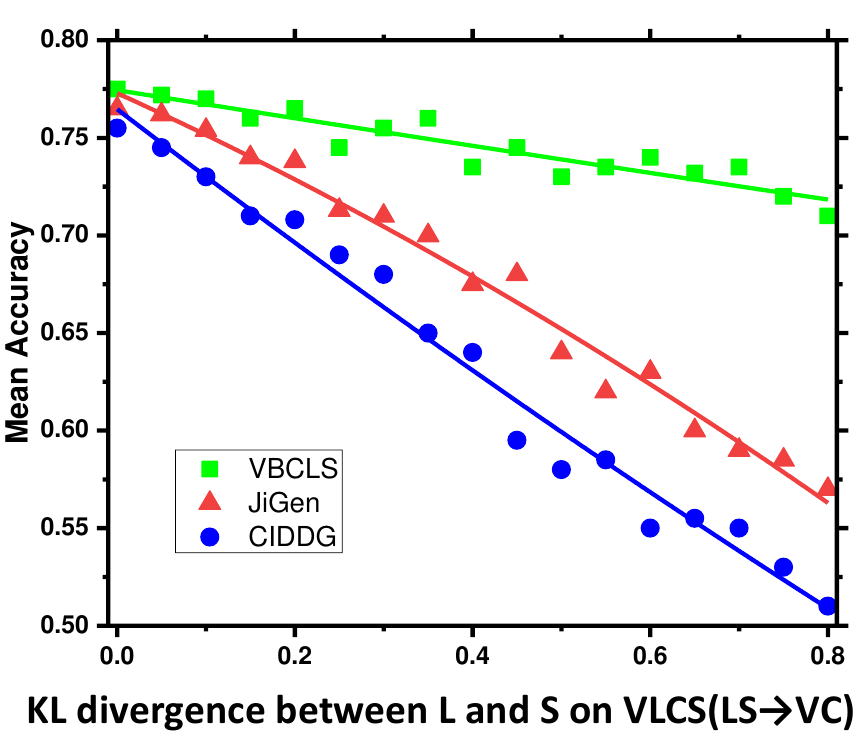}\\ 
\caption{Left: Quantification of DG performance on VLCS (LS$\rightarrow$VC) under different source label shifts ($p^L(y)\neq p^S(y)$ by sampling). }\label{fig:e1} 
\end{figure}


\section{Experiments}

This section demonstrates the effectiveness of our variational Bayesian inference framework under conditional and label shift (VBCLS) on the classic VLCS DG benchmark for image classification and the PACS benchmark for object recognition with domain shift. 

\subsection{Implementation Details}

The domain invariant encoder $f$ and posterior alignment classifier $f_{pa}$ use the encoder and classifier structure as our compared models (e.g., AlexNet and ResNet18), and the label prior network is simply the concatenation of $f$ and $f_{pa}$, and the likelihood maximization network $f_{lm}$ uses the reversed CNN decoder structure of $f$. We implement our methods using PyTorch, we empirically set $\alpha=0.5$, $\beta=1$, $\theta=0.1$ and $t=100$ via grid searching. In our experiments, the performance does not sensitive to these hyperparameters for a relatively large range.

Following previous work on domain generalization \cite{li2017deeper,li2019episodic}, we use models pre-trained on the ILSVRC dataset and label smoothing on the task classifier in order to prevent overfitting. The model is trained for 30 epochs via the mini-batch stochastic gradient descent (SGD) with a batch size of 128, a momentum of 0.9, and a weight decay of $5e-4$. The initial learning rate is set to $1e-3$, which is scaled by a factor of 0.1 after 80\% of the epochs. For the VLCS dataset, the initial learning rate is set to $1e-4$, since a high learning rate leads to early convergence due to the overfitting in the source domain. In addition, the learning rate of the domain discriminator and the classifier is set to be larger (i.e., 10 times) than that of the feature extractor.

For the ablation study, VBCLS-$f_{pa}$ denotes without posterior alignment and VBCLS-$L_{\hat{y}}$ denotes without minimizing $D_{KL}(q(z|x,y)||q(z|x,\hat{y}))$. 

\subsection{VLCS Dataset}

VLCS \cite{ghifary2016scatter} contains images from four different datasets including PASCAL VOC2007 (V), LabelMe (L), Caltech-101 (C), and SUN09 (S). Different from PACS, VLCS offers photo images taken under different camera types or composition bias. The domain V, L, C, and S have 3,376, 2,656, 1,415, and 3,282 instances, respectively. Five shared classes are collected to form the label space, including bird, car, chair, dog, and person. We follow the previous works to exploit the publicly available pre-extracted DeCAF6 features (4,096-dim vector) \cite{donahue2014decaf} for leave-one-domain-out validation by randomly splitting each domain into 70\% training and 30\% testing. We report the mean over five independent runs for our results.

The mean accuracy on the test partition using leave-one-domain-out validation on VLCS is reported in Table \ref{res:tab_vlcs}. We also compare our method with the DG models trained using the same architecture and source domains.

Our results indicate that our VBCLS outperforms the covariant shift setting methods (e.g., CCSA \cite{motiian2017unified}, MMD-AAE \cite{li2018domain}) by a large margin. The improvement over the conditional shift only method CIDDG \cite{li2018deepcc} is significant, demonstrating the necessity of incorporating both conditional and label shift. When compared with the recent SOTA methods, e.g., self-challenging based RSC \cite{huang2020self}, information bottleneck based MetaVIB \cite{du2020learning}, and self-training based EISNet \cite{wang2020learning}, we can observe that our VBCLS yields better performance in almost all cases. We note that JiGen \cite{carlucci2019domain} uses the Jigsaw Puzzles solving, which are essentially different from the IFL. 

\begin{table}[t]
\centering

\resizebox{1\columnwidth}{!}{
\begin{tabular}{l|cccc|c}
\hline
Target ($\rightarrow$) & V     & L     & C     & S     & Average \\ \hline\hline
CCSA 2017        & 67.10 & 62.10 & 92.30 & 59.10 & 70.15   \\
MMD-AAE 2018     & 67.70 & 62.60 & 94.40 & 64.40 & 72.28   \\
CIDDG 2018      & 64.38 & 63.06 & 88.83 & 62.10 & 69.59\\
Epi-FCR 2019     & 67.10 & 64.30 & 94.10 & 65.90 & 72.90   \\
JiGen 2019       & 70.62 & 60.90 & 96.93 & 64.30 & 73.19 \\
MetaVIB 2020      & 70.28& 62.66& 97.37& 67.85& 74.54\\
RCS 2020         &73.93 & 61.86 & 97.61 & 68.32 & 75.43\\

\hline\hline

VBCLS       & 72.16 & 68.63 & 96.52 & 70.37 & 76.92$\pm$0.06    \\ \hline

VBCLS-$f_{pa}$      & 69.40 & 65.00 & 94.60 & 65.60 & 73.70$\pm$0.08    \\
VBCLS-$L_{\hat{y}}$      & 71.74 & 68.18 & 96.20 & 70.02 & 76.56$\pm$0.07    \\\hline
\end{tabular}} \caption{Classification accuracy (mean$\pm$sd) using leave-one-domain-out validation on the VLCS benchmark.}
\label{res:tab_vlcs}
\end{table}

The good performance of our strategy indicates the invariant feature learning works well, and the conditional and shift assumption can be an inherent property that needs to be addressed for real-world DG challenges. The discrepancy between marginal distributions $\bar{p}(y)$ and $p^t(y)$ is measured via the KL-divergence as the semi-supervised learning with the selective bias problem \cite{zadrozny2004learning}. The impact of the label shift is empirically illustrated in Fig. \ref{fig:e1} left. In Fig. \ref{fig:e1} right, we show the label refinement can effectively estimate the testing label distribution without tuning the network. The label alignment can be relatively accurate after 3 epochs and is almost stable after 5 epochs.  

In the ablation study, we can see that posterior alignment is necessary if there is label shift.  Besides, the performance of the label prior network can be improved by minimizing $D_{KL}(q(z|x,y)||q(z|x,\hat{y}))$ on encoder $f$. 


\begin{table}[]
\centering
\resizebox{1\columnwidth}{!}{
\begin{tabular}{cc|cccccc}
\hline
\multicolumn{1}{l}{} & \multicolumn{1}{l}{} & \multicolumn{6}{|c}{Source}                    \\ \hline
Target               & Method               & VC    & VL    & VS    & LC    & LS    & CS    \\ \hline\hline
\multirow{2}{*}{V}   & CIDDG                   & -     & -     & -     & 60.42 & 62.21 & 59.56 \\
                     & VBCLS             & -     & -     & -     & 65.82 & 68.45 & 66.76 \\\hline
\multirow{2}{*}{L}   & CIDDG                   & 53.24 & -     & 52.27 & -     & -     & 49.58 \\
                     & VBCLS             & 60.75 & -     & 60.76 & -     & -     & 58.82 \\\hline
\multirow{2}{*}{C}   & CIDDG                   & -     & 78.82 & 78.58 & -     & 74.67 & -     \\
                     & VBCLS             & -     & 85.56 & 86.68 & -     & 81.47 & -     \\\hline
\multirow{2}{*}{S}   & CIDDG                   & 59.04 & 56.29 & -     & 59.80 & -     & -     \\
                     & VBCLS             & 62.34 & 60.35 & -     & 61.52 & -     & -     \\ \hline
\end{tabular}}\caption{Impact of reducing the number of source domains on VLCS. Note that the rows denote the two source domains.} 
\label{tab:source_div} 
\end{table}

Another evaluation protocol on VLCS is to examine whether eliminating examples from one source domain impacts the performance on the target domain. This protocol is designed to evaluate the impact of the diversity of the source domains on the target accuracy. In this experiment, each target domain on models by training all combinations of the remaining domains as the source domains is evaluated, as shown in Table \ref{tab:source_div}. In addition, the CIDDG baseline is included for reference. Results in Table \ref{tab:source_div} demonstrate that for all target domains, reducing the number of source domains from 3 (see Table \ref{res:tab_vlcs}) to 2 degrades the performance for all combinations of the source domains. We can see that, in some cases, excluding a particular source from the training substantially degrades the target loss. However, we can see that our VBCLS can still be more robust under these cases.




\begin{table}[t]
\resizebox{\columnwidth}{!}{
\begin{tabular}{l|ccccccc}
\hline
&Guitar &House& Giraffe& Person& Horse& Dog&Elephant\\\hline
Art Paint &184& 295 &285& 449& 201& 379& 255\\
Cartoon& 135& 288& 346 &405& 324 &389 &457\\
Photo& 186& 280& 182& 432 &199& 189 &202\\
Sketch &608& 80 &753 &160 &816 &772& 740\\
\hline
\end{tabular}
}\caption{Sample sizes for each $<$domain, class$>$ pair in PACS dataset. The column shows category name.}
\label{res:statpacs}
\end{table}

\subsection{PACS Dataset}

The object recognition benchmark PACS \cite{li2017deeper} consists of images divided into 7 classes from four different datasets, including Photo (P), Art painting (A), Cartoon (C), and Sketch (S). In Tab. \ref{res:statpacs}, we provide the detailed statistics of the number of samples in each domain. It is clear that the class ratio is different across domains, which indicates the significant label shift.

As shown in the results (c.f. Table \ref{res:tab_pacs}), the Sketch domain produces the lowest accuracy when used as the target domain, and therefore it is deemed the most challenging one. In light of this, we follow the previous work to tune the model using the S domain as the target domain and reuse the same hyperparameters for the experiments with the remaining domains. In Table \ref{res:tab_pacs}, we show the result using our method, which was averaged over 5 different initializations alongside all the other comparison methods. Overall, we can see that our method yields better average performance over all source domains compared to previous SOTA methods. Among them, CrossGrad \cite{shankar2018generalizing} synthesizes data for a new domain, MMLD \cite{matsuura2020domain} using a mixture of multiple latent domains. It is also comparable to the recent self-challenging based RSC \cite{huang2020self}, information bottleneck based MetaVIB \cite{du2020learning}, and self-training based EISNet \cite{wang2020learning}. More importantly, we can observe that our method outperforms CIDDG {\cite{li2018deepcc}}, an adversarial conditional IFL strategy, by a large margin. The ablation studies are also consistent with the results in VLCS and PACS. We also provide the results using the ResNet18 backbone.


\begin{table}[]
\centering

\resizebox{1\columnwidth}{!}{
\begin{tabular}{l|cccc|c}
\hline
Target ($\rightarrow$)     & P     & A     & C     & S     & Average \\ \hline\hline

CrossGrad 2018  & 87.6 & 61.0   & 67.2  & 55.9 & 67.9   \\

CIDDG 2018    & 78.7  & 62.7   & 69.7  & 64.5 & 68.9   \\

Epi-FCR 2019    & 86.1  & 64.7  & 72.3  & 65.0 & 72.0      \\
JiGen 2019      & 89.0  & 67.6 & 71.7 & 65.2 & 73.4   \\

MMLD 2020      & 88.98 &  69.27 & 72.83& 66.44& 74.38\\


RSC 2020 & 90.88 &71.62 & 66.62& 75.11 &76.05\\
EISNet 2020 & 91.20& 70.38&71.59 &70.25& 75.86\\
MetaVIB 2020 & 91.93 & 71.94  &73.17  &65.94  & 75.74\\\hline\hline

VBCLS   & 92.12 & 70.60  & 77.36 & 70.19 & 77.55$\pm$0.07    \\ \hline
VBCLS-$f_{pa}$ & 91.02 & 68.86 & 74.18 & 65.40  & 74.61$\pm$0.04    \\ 
VBCLS-$L_{\hat{y}}$ & 91.77 & 70.54 & 76.56  & 70.33 & 77.19$\pm$0.06   \\

JiGen 2019 [Res18]   &96.03 &79.42 &75.25 &71.35 &80.51   \\ 
MMLD 2020 [Res18]   &96.09  & 81.28& 77.16& 72.29 & 81.83  \\
EISNet 2020 [Res18]& 95.93& 81.89&76.44 &74.33& 82.15\\
RSC 2020  [Res18]&95.99&83.43& 80.85  &80.31 &85.15\\\hline

VBCLS [Res18]   & 97.21 & 84.63  & 82.06 & 79.25 & 86.73$\pm$0.05    \\ \hline

\end{tabular}
}\caption{Classification accuracy (mean$\pm$sd) using leave-one-domain-out validation on PACS.}
\label{res:tab_pacs}
\end{table}

\section{Conclusion}\label{sec:conc}


In this work, we target to establish a more realistic assumption in DG that both conditional and label shifts arise independently and contemporarily. We theoretically analyze the inequilibrium of conventional IFL under the different shift assumptions. Motivated by that, a concise yet effective VBCLS framework based on variational Bayesian inference with the posterior alignment is proposed to reduce both the conditional shift and label shifts. Extensive evaluations verify our analysis and demonstrate the effectiveness of our approach, compared with IFL approaches.



\section*{Acknowledgments}
This work was supported in part by the Hong Kong GRF 152202/14E, PolyU Central Research Grant G-YBJW, and Jiangsu NSF (BK20200238). 

\bibliographystyle{named}
\bibliography{ijcai21}

\end{document}